\documentclass[11pt,a4paper]{article}
\usepackage[nohyperref]{naaclhlt2019}
\usepackage{times}
\usepackage{latexsym}
\usepackage{graphicx}  
\usepackage{tabularx}
\usepackage{multirow}
\usepackage{multicol}
\usepackage{url}
\usepackage{xcolor}
\aclfinalcopy 

\title{Predicting Algorithm Classes for Programming Word Problems}
\author{Vinayak Athavale\thanks{equal contribution}, Aayush Naik \footnotemark[1], Rajas Vanjape, Manish Shrivastava \\
IIIT Hyderabad \\
vinayak.athavale@research.iiit.ac.in \\
}
\date{}

\begin{document}
\maketitle
\begin{abstract}
  We introduce the task of algorithm class prediction for programming word problems. A programming word problem is a problem written in natural language, which can be solved using an algorithm or a program. We define classes of various programming word problems which correspond to the class of algorithms required to solve the problem. We present four new datasets for this task, two multiclass datasets with 550 and 1159 problems each and two multilabel datasets having 3737 and 3960 problems each. We pose the problem as a text classification problem and train neural network and non-neural network based models on this task. Our best performing classifier gets an accuracy of 62.7 percent for the multiclass case on the five class classification dataset, Codeforces Multiclass-5 (CFMC5). We also do some human-level analysis and compare human performance with that of our text classification models. Our best classifier has an accuracy only 9 percent lower than that of a human on this task. To the best of our knowledge, these are the first reported results on such a task. We make our code and datasets publicly available. 

\end{abstract}

\begin{figure}

  \begin{minipage}{0.48\textwidth}
  \centering
      \includegraphics[scale=.39]{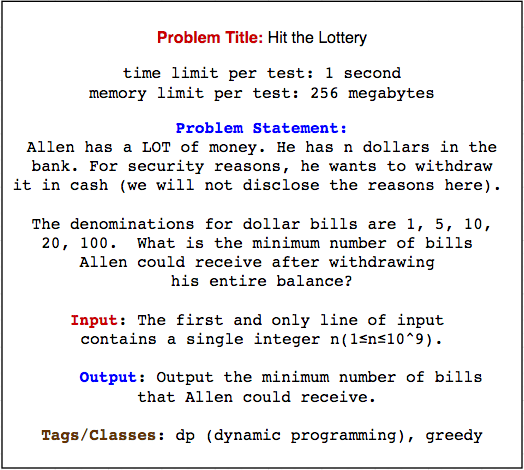}
      \caption{An example programming word problem. Note that the example shown here is one of the easy Codeforces problems -- most problems are much harder.}
      \label{fig:example}
  \end{minipage}
  
\end{figure}

\section{Introduction}

\label{sec:in}

In this paper we introduce and work on the problem of predicting algorithms classes for programming word problems (PWPs). A PWP is a problem written in natural language which can be solved using a computer program. These problems generally map to one or more classes of algorithms, which are used to solve them. Binary search, disjoint-set union, and dynamic programming are some examples. In this paper, our aim is to automatically map programming word problems to the relevant classes of algorithms. We approach this problem by treating it as a classification task.

{\bf Programming word problems}
A programming word problem (PWP) requires the solver to design correct and efficient programs. The correctness and efficiency is checked by various test-cases provided by the problem writer. A PWP usually consists of three parts -- the problem statement, a well-defined input and output format, and time and memory constraints. An example PWP can be seen in Figure \ref{fig:example}.

Solving PWPs is difficult for several reasons. One reason is, the problems are often embedded in a narrative, that is, they are described as quasi real-world situations in the form of short stories or riddles. The solver must first decode the intent of the problem, or understand \emph{what} the problem is. Then the solver needs to apply their knowledge of algorithms to write a solution program. Another reason is that, the solution programs must be efficient with respect to the given time and memory constraints. An outgrowth of this is that, the algorithm required to solve a particular problem not only depends on the problem statement, but also the constraints. Consider that there may be two different algorithms which will generate the correct output, for example, linear search, and binary search, but only one of those will abide by the time and memory constraints. With the growing popularity of these problems, various competitions like ACM-ICPC, and Google CodeJam have emerged. Additionally, several companies including Google, Facebook, and Amazon evaluate problem-solving skills of candidates for software-related jobs \citep{mcdowell2016cracking} using PWPs. Consequently, as noted by \citet{forivsek2010difficulty}, programming problems have been becoming more difficult over time. To solve a PWP, humans get information from all its parts, not just the the problem statement. Thus, we predict algorithms from the entire text of a PWP. We also try to identify which parts of a PWP contribute the most towards predicting algorithms.


{\bf Significance of the Problem}
Many interesting real-world problems can be solved and optimised using standard algorithms. Time spent grocery shopping can be optimised by posing it as a graph traversal problem \cite{gertin2012maximizing}. Arranging and retrieving items like mail, or books in a library can be done more efficiently using sorting and searching algorithms. Solving problems using algorithms can be scaled by using computers, transforming the algorithms into programs.  A program is an algorithm that has been customised to solve a specific task under a specific set of circumstances using a specific language. Converting textual descriptions of such real-world problems into algorithms, and then into programs has largely been a human endeavour. An AI agent that could automatically generate programs from natural language problem descriptions could greatly increase the rate of technological advancement by quickly providing efficient solutions to the said real-world problems. A subsystem that could identify algorithm classes from natural language would significantly narrow down the search space of possible programs. Consequently, such a subsystem would be a useful feature for, or likely be even part of, such an agent. Therefore, building a system to predict algorithms from programming word problems is potentially an important first step toward an automatic program generating AI.
More immediately, such a system could serve as an application to help people in improving their algorithmic problem-solving skills for software job interviews, competitive programming, and other uses.

As per our knowledge, this task has not been addressed in the literature before. Hence, there is no standard dataset available for this task. We generate and introduce new datasets by extracting problems from Codeforces\footnote{codeforces.com}, a sport programming platform. We release the datasets and our experiment code at $masked$\footnote{hidden for the the double blind review}.

{\bf Contribution} The major contributions of this paper are: {\bf Four datasets} on programming word problems - two multiclass\footnote{each problem belongs to only one class} datasets having 5 and 10 classes and two multilabel\footnote{each problem belongs to one or more classes} datasets having 10 and 20 classes.
{\bf Evaluation of Classifiers} on various multiclass and multilabel classifiers that can predict classes for programming word problems on our datasets along with the human baseline. We define our problem more clearly in section \ref{sec:pd}. Then we explain our datasets -- their generation and format along with human evaluation in section \ref{sec:ds}. We describe the models we use for multiclass and multilabel classification in section \ref{sec:cm}. We delineate our experiments, models, and evaluation metrics in section \ref{sec:es}. We report our classification results in section \ref{sec:rs}. We analyse some dataset nuances in section \ref{sec:an}. Finally, we discuss related work and the conclusion in sections \ref{sec:rw} and \ref{sec:fw} respectively.

\begin{table*}[t]
\label{table:1}
\begin{tabularx}{\textwidth}{|l|l|l|l|l|X|}
\hline
 \bf Dataset & \bf Size & \bf Vocab & \bf classes  & \bf Avg. words & \bf Class percentage\\  \hline 
CFMC5 & 550 & 9326 & 5 & 504 & {\color{red} greedy: 20\%}, {\color{red}implementation:20\%},
 {\color{blue} data structures: 20\%}, {\color{red} dp: 20\%}, {\color{blue} math: 20\%}  \\  \hline 
CFMC10 & 1159 & 14691 & 10 & 485 & {\color{red}implementation: 34.94\%}, {\color{red}dp: 12.42\%},
 {\color{blue}math: 11.38\%},  {\color{red}greedy: 10.44\%}, {\color{blue}data structures: 9.49\%}, {\color{red}brute force: 5.60\%}, {\color{blue}geometry: 4.22\%},  
 {\color{red}constructive algorithms: 5.52\%}, 
  {\color{red}dfs and similar: 3.10\%}, {\color{blue}strings: 2.84\%}  \\ \hline 

\end{tabularx}
\caption{Dataset statistics for multiclass datasets. CFMC5 has 550 problems with a balanced class distribution. CFMC10 has 1159 problems and has a class imbalance. CFMC5 is a subset of CFMC10.  {\color{red}Red} classes belong to the solution category; {\color{blue}blue} classes belong to the problem category.}\label{table:multiclassstat}
\end{table*}

\begin{table*}[t]
\label{table:2}
\begin{tabularx}{\textwidth}{|l|l|l|l|l|l|l|X|}
\hline  \bf Dataset & \bf Size & \bf Vocab & \bf N classes  & \bf  Avg. len & \bf Label card & \bf Label den &  \bf Label subsets \\  
\hline
CFML10 & 3737 & 28178 & 10 & 494 & 1.69  & 0.169 & 231\\ 
CFML20 & 3960 & 29433 & 20 & 495 & 2.1 & 0.105 & 808\\ \hline

\end{tabularx}
\caption{Dataset statistics for multilabel datasets. The problems of the CFML10 dataset are a subset of those in the CFML20 dataset.} \label{table:multilabelstat}
\end{table*}

\section{Problem Definition}
\label{sec:pd}
The focus of this paper is the problem of mapping a PWP to one or more classes of algorithms. A \emph{class} of algorithms is a set containing more specific algorithms. For example, breadth-first search, and Dijkstra's algorithm belong to the class of graph algorithms. A PWP can be solved using one of the algorithms in the class it is mapped to. Problems on the Codeforces platform have \emph{tags} that correspond to the class of algorithms.

Thus, our aim is to find a tagging function, $f^* : \mathcal{S} \rightarrow \mathcal{P}(\mathcal{T})$ which maps a PWP string, $s \in \mathcal{S}$, to a set of tags, $\{t_1, t_2, ...\} \in \mathcal{P}(\mathcal{T})$.
We also consider another variant of the problem. For the PWPs that only have one tag, we focus on finding a tagging function, $f_1^* : \mathcal{S} \rightarrow \mathcal{T}$, which maps a PWP string, $s \in \mathcal{S}$, to a tag, $t \in \mathcal{T}$. We approximate $f^*$ and $f_1^*$ by training models on data.

\section{Dataset}
\label{sec:ds}

\subsection{Data Collection}
We collected the data from a popular sport programming platform called Codeforces. Codeforces was founded in 2010, and now has over 43000 active registered participants\footnote{http://codeforces.com/ratings/page/219}.
We first collected a total of 4300 problems from this platform. Each problem has associated tags, with most of the problems having more than one tag. These tags correspond to the algorithm or class of algorithms that can be used to solve that particular problem. The tags for a problem are given by the problem writer and they can only be edited only by high-rated (expert) contestants who have solved the problem.
Next, we performed basic filtering on the data -- removing the problems which had non-algorithmic tags, problems with no tags assigned to them, and also the problems wherein the problem statement was not extracted completely. After this filtering, we got 4019 problems with 35 different tags. This forms the Codeforces dataset. The label (tag) cardinality (average number of labels/tags per problem) was 2.24. 
Since the Codeforces dataset is the first dataset generated for a new problem, we select different subsets of this dataset with differing properties. This is to check if classification models are robust to different variations of the problem.

\subsection{Multilabel Datasets}
We found that a large number of tags had a very low frequency. Hence, we removed those problems and tags from the Codeforces dataset as follows. First, we got the list of 20 most frequently occurring tags, ordered by decreasing frequency. We observed that the $20^{th}$ tag in this list had a frequency of 98, in other words, 98 problems had this tag. Next, for each problem, we removed the tags that are not in this list. After that, all problems that did not have any tags left were removed.

This led to the formation of the Codeforces Multilabel-20 (CFML20) dataset, which has 20 tags. We used the same procedure for the 10 most frequently occurring tags to get the Codeforces Multilabel-10 (CFML10) dataset. The CFML20 has 98.53 (3960 problems) percent of the problems of the original dataset and the label (tag) cardinality only reduces from 2.24 to 2.21. CFML10 on the other hand has 92.9 percent of the problems with label (tag) cardinality 1.69. Statistics about both these multilabel datasets are given in Table \ref{table:multilabelstat}.

\subsection{Multiclass Datasets}
To generate the multiclass datasets, first, we extracted the problems from the CFML20 dataset that only had one tag. There were about 1300 such problems. From those, we selected the problems whose tags occur in the list of 10 most common tags. These problems formed the Codeforces Multiclass-10 (CFMC10) dataset which contains 1159 examples. We found that the CFMC10 dataset has a class (tag) imbalance.
We also make a balanced dataset, Codeforces Multiclass-5 (CFMC5), in which the prior class (tag) distribution is uniform. The CFMC5 dataset has five tags, each having 110 problems. To make CFMC5, first we extracted the problems whose tags are among the five most common tags. The fifth most common tag occurs 110 times. We sampled 110 random problems corresponding to the other four tags to give a total of 550 problems. Statistics about both the multiclass datasets are given in Table \ref{table:multiclassstat}.

\subsection{Dataset Format}
Each problem in the datasets follows the same format (refer to Figure \ref{fig:example} for an example problem). The header contains the problem title, and the time and memory constraints for a program running on the problem testcases. The problem statement is the natural language description of the problem framed as a real world scenario.
The input and output format describe the input to, and the output from a valid solution program. It also contains constraints that will be put on the size of inputs (for example, max size of input array, max size of 2 input values). 
The tags associated with the problem are the algorithm classes that we are trying to predict using the above information.


\subsection{Class Categories in the Dataset}
The classes for PWPs can be divided into two categories:  
{\bf Problem category} classes tell us what kind of broad class of problem the PWP belongs to. For instance, \emph{math}, and \emph{string} are two such classes. 
{\bf Solution category} classes tell us what kind of algorithm can solve a particular PWP. For example, a PWP of class \emph{dp} or \emph{binary search} would need a dynamic programming or binary search based algorithm to solve it. 

Problem category PWPs are easier to classify because, in some cases, simple keyword mapping may lead to the classification (an equation in the problem is a strong indicator that a problem is of math type). Whereas, for solution category PWPs, a deeper understanding of the problem is required.

The classes belong to problem and solution categories for CFML20 are mentioned in the supplementary material.


\subsection{Human Evaluation}
In this section, we evaluate and analyze the performance of an average competitor on the task of predicting an algorithm for a PWP. The tags for a given PWP are added by its problem setter or other high-rated contestants who have solved it. Our test participants were recent computer science graduates with some experience in algorithms and competitive programming. We gave 5 participants the problem text along with all the constraints, and the input and output format. We also provided them with a list of all the tags and a few example problems for each tag. We randomly sample 120 problems from the CFML20 dataset and split them into two parts -- containing 20 and 100 problems respectively. The 20 problems were given along with their tags to familiarize the participants with the task. For the remaining 100 problems, the participants were asked to predict the tags (one or more) for each problem. We chose to sample the problems from the CFML20 dataset as it is the closest to a real-world scenario of predicting algorithms for solving problems. We find that there is some variation in the accuracy reported by different humans with the highest F1 micro score being 11 percent greater than that of the the lowest. (see supplementary material for more details). The F1 micro score averaged over all 5 participants was 51.8 while the averaged F1 Macro was 42.7. The results are not surprising since this task is like any other problem solving task, and people based on their proficiency would get different results. This shows us that the problem is hard even for humans with a computer science education.



\section{Classification Models}
\label{sec:cm}
To test the compatibility of our problem with text classification paradigm, we apply to it some standard text classification models from recent literature.

\subsection{Multiclass Classification}
To approximate the optimal tagging function $f_1^*$ (see section \ref{sec:pd}) we use the following models.

{\bf Multinomial Naive Bayes (MNB) and Support Vector Machine (SVM)}
\citet{Wang:2012:BBS:2390665.2390688}  proposed  several  simple and effective  baselines  for  text  classification. An MNB is a naive Bayes classifier for multinomial models. An SVM  is a discriminative hyperplane-based classifier \cite{hearst1998support}. These baselines use unigrams and  bigrams as features. We also try applying TF-IDF to these features.

{\bf Multi-layer Perceptron (MLP)}
An MLP is a class of artificial neural network that uses backpropagation for training in a supervised setting \cite{rumelhart:errorpropnonote}. MLP-based models are standard for text classification baselines \cite{glorot2011domain}.

{\bf Convolutional Neural Network (CNN)}
We also train a Convolutional Neural Network (CNN) based model, similar to the one used by  \citet{kim2014convolutional} in their paper, to classify the problems.  We use the model both with and without pre-trained GloVe word-embeddings \cite{Pennington14glove:global}.

{\bf CNN ensemble}
\citet{58871} introduce neural network ensemble learning, in which many neural networks are trained and their predictions combined. These neural network systems show greater generalization ability and predictive power. We train five CNN networks and combine their predictions using the majority voting system.

\subsection{Multilabel Classifiers}
To approximate, $f^*$ (see section \ref{sec:pd}), we apply the following augmentations to the models described above.

{\bf Multinomial Naive Bayes (MNB) and Support Vector Machine (SVM)}
For applying these models to the multilabel case, we use the one-vs-rest (or, one-vs-all) strategy. This strategy involves training a single classifier for each class, with the samples of that class as positive samples and all other samples as negatives \cite{Bishop:2006:PRM:1162264}.

{\bf Multi-layer Perceptron (MLP)}
\citet{nam2014large} use MLP-based models for multilabel text classification. We use similar models, but use the MSE loss instead of the cross-entropy loss.

{\bf Convolutional Neural Network (CNN)}
For multilabel classification we use a CNN based feature extractor similar to the one used in \cite{kim2014convolutional}. The output is passed through a sigmoid activation function, $\sigma(x) = \frac{1}{1 + e^{-x}}$. The labels which have a corresponding activation greater than 0.5 are considered \cite{liu2017deep}. Similar to the multiclass case, we train the model both with and without pre-trained GloVe \cite{Pennington14glove:global} word-embeddings. 

{\bf CNN ensemble}
We train five CNNs and add their output linear activation values. We pass this sum through a sigmoid function and consider the labels (tags) with activation greater than 0.5.
\section{Experiment setup} \label{sec:es}
All hyperparameter tuning experiments were performed with 10-fold cross validation. For the non-neural network-based methods, we first vectorize each problem using a bag-of-words vectorizer, scikit-learn's \cite{scikit-learn} CountVectorizer. We also experiment with TF-IDF features for each problem.  
In the multiclass case, we use the LIBSVM \cite{Chang01libsvm:a} implementation of the SVM classifier and we grid search over different kernels. However, the LIBSVM implementation is not compatible with the one-vs-rest strategy (complexity $\mathcal{O}(n)$ where $n$ is the number of classes), but only the one-vs-one (complexity $\mathcal{O}(n^2)$). This becomes prohibitively slow and thus, we use the LIBLINEAR \cite{REF08a} implementation for the multilabel case. For hyperparameter tuning, we applied a grid search over the parameters of the vectorizers, classifiers, and other components. The exact parameters tuned can be seen in our code repository.
For the neural network-based methods, we tokenize each problem using the spaCy tokenizer \cite{spacy2}.  We only use words appearing 2 or more times in building the vocabulary and replace the words that appear fewer times with a special UNK token.  Our CNN network architecture is similar to that used by \citet{kim2014convolutional}. The batch size used is 32. We apply 512 one-dimensional convolution filters of size 3, 4, and 5 on each problem. The rectifier, $R(x) = max(x, 0)$, is used as the activation function. We concatenate these filters, apply a global max-pooling followed by a fully-connected layer with output size equal to the number of classes. We use the PyTorch framework \cite{paszke2017automatic} to build this model. For the word embedding we use two approaches - a vanilla PyTorch trainable embedding layer and a 300-dimensional GloVe embedding \cite{Pennington14glove:global}. The networks were initialized using the Xavier method \cite{Glorot10understandingthe} at the beginning of each fold. We use the Adam optimization algorithm \cite{kingma2014adam} as we observe that it converges faster than vanilla stochastic gradient descent.


\section{Results}
\label{sec:rs}
\begin{table}[t!]
\begin{tabularx}{\textwidth}{|l|l|l|l|l|}
 \cline{1-5}  \multirow{2}{*}{\bf Classifier}
      & \multicolumn{2}{c}{\bf CFMC5}
          & \multicolumn{2}{c|}{\bf CFMC10} \\ \cline{2-5}             
  & Acc & F1 W & Acc & F1 W \\  \cline{1-5}  
CNN Random & 25.0 & 22.1 & 35.2 & 19.2 \\
MNB & 47.6 & 47.5 & 43.9 & 37.4 \\
SVM BoW & 49.3  & 49.1 & 47.9 & 43.2 \\
SVM TFIDF & 47.8 & 47.6 & 45.7 & 41.2  \\
MLP & 47.8 & 47.6 & 49.3 & 46.2  \\
CNN & 61.7 & 61.3 & \bf{54.7} & 51.3 \\
CNN Ensemble & \bf{62.7} & \bf{62.2} & 53.5 & 50.5 \\
CNN GloVe &  62.2 & 61.3 & 54.5 & \bf{51.4} \\
\cline{1-5}
\end{tabularx} 

\caption{Classification Accuracy for single label classification. Note that all results were obtained on 10-fold cross validation. CNN Random refers to a CNN trained on a random labelling of the dataset. F1 W stands for weighted macro F1-score.}\label{table:multiclass}
\end{table}

\begin{table*}[t]
\begin{tabularx}{\textwidth}{|l|l|l|l|l|l|l|}
\cline{1-7}   \multirow{2}{*}{\bf Classifier}
      & \multicolumn{3}{c}{\bf CFML10}
          & \multicolumn{3}{c|}{\bf CFML20} \\ \cline{2-7}            
 &  hamming loss & F1  micro & F1  macro & hamming loss & F1 micro & F1  macro \\  \cline{1-7}
CNN Random TWE & 0.2158 & 15.98 & 9.39 & 0.1207 & 12.07 & 4.02 \\
MNB BoW & 0.1706 & 30.57 & 25.73 & 0.1067 & 29.67 & 23.41 \\
SVM BoW & 0.1713 & 36.08 & 31.09 & 0.1056 & 34.93 & 30.70 \\
SVM BoW + TF-IDF & 0.1723 & 38.20 & 33.68 & 0.1059 & 38.55 & 34.70 \\
MLP BoW & 0.1879 & 39.13 & 34.92 & 0.1167 & 38.12 & 31.37\\
CNN TWE  &  \bf{0.1671} & 39.20 & 32.59 & 0.\bf{1023} & 38.44 & 30.38 \\
CNN Ensemble TWE & 0.1703 & \bf{45.32} & \bf{38.93} & 0.1093 & \bf{42.75} & \bf{37.29} \\
CNN  GloVe & 0.1676 & 39.22 & 33.77 & 0.1052 & 37.56 & 30.29 \\
Human &- &- &- & - & \bf{51.8} & \bf{42.7} \\
\cline{1-7}

\end{tabularx}
\caption{Classification Accuracy for multi-label classification. TWE stands for trainable word embeddings initialized with a normal distribution. Note that all results were obtained on 10-fold cross validation. CNN Random refers to a CNN trained on a random labelling of the dataset. }\label{table:multilabel}
\end{table*}

\subsection{Multiclass Results}

We see that the classification accuracy of the best performing classifier, CNN ensemble, for the CFMC5 dataset is 62.7 \%. The highest accuracy for the CFMC10 dataset was achieved by the CNN classifer which does not use any pretrained embeddings.  For all the multiclass classification results refer to table \ref{table:multiclass}.  We observe that CNN-based classifiers perform better than other classifiers -- MLP, MNB, and SVM for both CFMC5 and CFMC10 datasets. Since these are the first learning results on the task of algorithm prediction for PWPs, we train a CNN classifier on a random labelling of the dataset. The results are given in the row called CNN random. To obtain this random labelling we shuffle the current mapping from problem to tag randomly. This ensures that the class distribution of the datasets remain the same. We see that all the classifiers significantly outperform the performance on the random dataset. 
We also observe that the classification accuracy is not the same for every class. We get the highest accuracy (see Fig. \ref{fig:confmat}) for the class, \emph{data structures}, at 90\%, while, the lowest accuracy is for the class, \emph{greedy}, at 40\%. These results are on the CFMC5 dataset.

\subsection{Multilabel Results}

We see that CNN-based classifiers give the best results for the CFML10 and CFML20 datasets. The best F1 micro and macro scores for the CFML10 dataset were 45.32, 38.9 respectively. These were obtained by the CNN Ensemble model. For complete results see table \ref{table:multilabel}. 
The best performing model on the CFML20 dataset was also the CNN ensemble. As we did in the multiclass case, we train a CNN model on the randomly shuffled labelling for both CFML10, CFML20 datasets. We find that all the classifers significantly outperform the model trained on a shuffled labelling.  
The human-level F1 micro and macro scores on a subset of the CFML20 dataset were 51.2 and 40.5. In comparison, our best performing classifier on the CMFL20 dataset, CNN Ensemble, got F1 macro and micro scores of 42.75, 37.29 respectively. We see that the performance of our best classifiers trail average human performance by about 8.45\% and 3.21\% on F1 micro and F1 macro scores respectively.

\section{Analysis}
\label{sec:an}

\subsection{Experiments with various subsets of the problem}

As described in section \ref{sec:in}, a PWP consists of three components -- the problem statement, input and output format, and time and memory constraints. We seek to answer the following questions. Does one component contribute to the accuracy more than any other? Does the contribution of different components vary over the problem class? We performed some experiments to address these questions.  We split the problem into two parts -- 1) the problem statement, and 2) the input and output format, and time and memory constraints. We train an SVM, and a CNN on these two components independently.

\begin{figure*}
 \center
  \includegraphics[width=\textwidth]{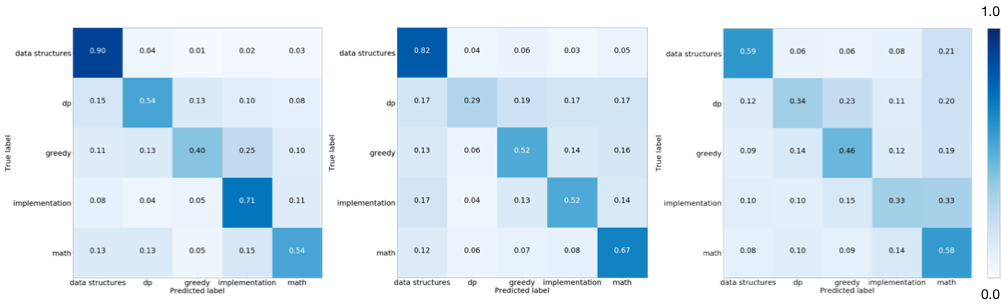}
  \caption{Confusion matrices for different parts of the problem on CFMC5. Whole problem text (left), only format and constraints information (center), and only problem statement (right). Perfomance on the whole problem is the highest, followed by only format and constraints information. Performance across different classes (except \emph{greedy}) is additive, which shows that features extracted from both the parts are of importance}\label{fig:confmat}
\end{figure*}

{\bf Multiclass PWP component analysis}
We find classifier accuracies on the CFMC5 dataset. We choose the CFMC5 dataset out of the two multiclass datasets because it has a balanced class distribution. We find that the classifiers perform quite well on only the input and output format, and time and memory constraints -- the best classifier getting an accuracy of 56.4 percent (only 5.3 percent lower than the accuracy of CNN with the whole problem). Classification using only the problem statement gives worse results than using the format and constraints, with a classification accuracy of 45.2 percent for the best classifier CNN (16.5 percent lower than the accuracy of a CNN trained on the whole problem). Complete results are given in table \ref{table:component}. 
We also see that the performance across different classes varies when trained on different inputs. We find that the class \emph{dp} performs better when trained on the problem statement, whereas the other classes perform much better on the format and constraints. For each class except \emph{greedy}, we see an additive trend -- the accuracy is improved by combining both these features. Refer to figure \ref{fig:confmat} for more details.
\begin{table*}[t]
\begin{tabularx}{\textwidth}{|l|l|l|X|X|X|X|X|X|}
\cline{1-9}   \multirow{2}{*}{\bf Dataset} &
              \multirow{2}{*}{\bf Features}&
              \multirow{2}{*}{\bf Classifier} 
      & \multicolumn{2}{c|}{\bf Soln. category}
      & \multicolumn{2}{c|}{\bf Prob. category}
          & \multicolumn{2}{c|}{\bf all} \\ \cline{4-9}            
 &   & & F1 Mi & F1 Ma  & F1 Mi  & F1 Ma & F1 Mi  & F1 Ma \\  \cline{1-9}
CFMC5 & only statement  & cnn  & 42.73 & 46.14 & 51.32 & 64.35 &  46.13  & 45.20 \\
CFMC5 & only i/o  & cnn  & 44.24 & 51.73 & 74.73 & 81.31 &  56.42 & 55.41 \\
CFMC5 & all prob  & cnn  & 54.24 & 59.91 & 71.36 & 78.32 & 61.71 & 61.32 \\
CFML20 & only statement  & cnn  & 30.83 & 17.32 & 38.64 & 41.82 & 33.59 & 28.34 \\
CFML20 & only i/o  & cnn  & 34.63 & 19.59 & 44.49 & 44.34 & 38.44 & 30.38 \\
CFML20 & all prob  & cnn  & 34.39 & 19.23 & 45.36 & 44.02 & 39.20 & 32.59 \\
\cline{1-9}
\end{tabularx}
\caption{Performance on different categories of PWPs for different parts of the PWPs. The rows with "only statement" features use only the problem description part of the PWP, the rows with "only i/o" use only the I/O and constraint information, and "all prob" use the entire PWP. The results under the "Soln category" column are of those problems that belong to the solution category, those under "Prob category" belong to the problem category, and those under "all" are for all the PWPs. So, for example, the F1 Micro score using only I/O and constraint for solution category problems of CFML20 is 34.63. Note that for CFMC5, F1 Mi (F1 Micro) is the same as accuracy, and F1 Ma (F1 Macro) score is a weighted Macro F1-score.}\label{table:component}
\end{table*}

{\bf Multilabel partial problem results}
We also tabulate the classifier accuracies on the CFML20 dataset by training it only on the format and constraints, and the problem statement. Even here, we observe similar trends as the multiclass partial problem experiments. We find that classifiers are more accurate when trained only on the format and constraints than only on the problem statement. Again, the accuracy is improved by combining both these features. Refer to table \ref{table:component} for more details.

\subsection{Problem category and Solution category results}
We find that correctly classifying PWPs of the solution category is harder than correctly classifying PWPs of the problem category (table \ref{table:component}). For instance, take a look at the row corresponding to CFMC5 dataset and "all prob" feature. The accuracy for solution category is 54.24\% as compared to 71.36\% for the problem category. This trend is followed for both CFMC5 and CFML20 datasets and also when using different features of the PWPs. In spite of the difficulty, the classification scores for the solution category are significantly better than random.


\section{Related Work}
\label{sec:rw}
Our work is related to three major topics of research, math word problem solving, text document classification and program synthesis.

{\bf Math word problem solving} 
In the recent years, many models have been built to solve different kinds of math word problems. Some models solve only arithmetic problems \cite{hosseini2014learning},  while others solve algebra word problems \cite{kushman2014learning}. There are some recent solvers which solve a wide range pre-university level math word problems \cite{matsuzaki2017semantic}, \cite{hopkins2017beyond}. \citet{wang2017deep}, and \citet{mehta2017deep} have built deep neural network based solvers for math word problems.
{\bf Program synthesis}
Work related to the task of converting natural language description to code comes under the research areas of program synthesis and natural language understanding. This work is still in its nascent stage.  \citet{zhong2017seq2sql} worked on generating SQL queries automatically from natural language descriptions. \citet{lin2017program} worked on automatically generating bash commands from natural language descriptions. \citet{iyer2016summarizing} worked on summarizing source code.
\citet{sudha2017classification} use a CNN based model to classify the algorithm used in a programming problem using the C++ code. Our model tries to accomplish this task by using the natural language problem description. 
\citet{gulwani2017program} is a comprehensive treatise on program synthesis.
{\bf Document classification} 
The problem of classifying a programming word problem in natural language is similar to the task of document classification. The state-of-the-art approach currently for single label classification is to use a hierarchical attention network based model \citep{yang2016hierarchical}. This model is improved by using transfer learning \cite{howard2018universal}. Other approaches include a Recurrent Convolutional Neural Network based approach \cite{lai2015recurrent} or the fasttext model \cite{joulin2016bag} which uses bag-of-words features and a hierarchical softmax. \citet{nam2014large} use a feed-forward neural network with binary cross entropy per label to perform multilabel document classification.  \citet{kurata2016improved} leverage label co-occurrence to improve multilabel classification.  \citet{liu2017deep} use a CNN based architecture to perform extreme multilabel classification.

\section{Conclusion}
\label{sec:fw}
We introduced a new problem of predicting the algorithm classes for programming word problems. For this task we generated four datasets -- two multiclass (CFMC5 and CFMC10), having five and 10 classes respectively, and two multilabel (CFML10 and CFML20), having 10 and 20 classes respectively.
Our classifiers are falling short only by about 9 percent of the human score. We also did some experiments which show that increasing the size of the train dataset improves the accuracy (see supplementary material). 
These problems are much harder than high school math word problems as they require a good knowledge of various computer science algorithms and an ability to reduce a problem to these known algorithms. Even our human analysis shows that trained computer science graduates only get an F1 of 51.8.
Based on these results, we see that algorithm class prediction is compatible with and can be solved using text classification. 

\bibliography{naaclhlt2019}

\begin{thebibliography}{36}
\expandafter\ifx\csname natexlab\endcsname\relax\def\natexlab#1{#1}\fi

\bibitem[{Bishop(2006)}]{Bishop:2006:PRM:1162264}
Christopher~M. Bishop. 2006.
\newblock \emph{Pattern Recognition and Machine Learning (Information Science
  and Statistics)}.
\newblock Springer-Verlag, Berlin, Heidelberg.

\bibitem[{chung Chang and Lin(2001)}]{Chang01libsvm:a}
Chih chung Chang and Chih-Jen Lin. 2001.
\newblock Libsvm: a library for support vector machines.

\bibitem[{Fan et~al.(2008)Fan, Chang, Hsieh, Wang, and Lin}]{REF08a}
Rong-En Fan, Kai-Wei Chang, Cho-Jui Hsieh, Xiang-Rui Wang, and Chih-Jen Lin.
  2008.
\newblock {LIBLINEAR}: A library for large linear classification.
\newblock \emph{Journal of Machine Learning Research}, 9:1871--1874.

\bibitem[{Fori{\v{s}}ek(2010)}]{forivsek2010difficulty}
Michal Fori{\v{s}}ek. 2010.
\newblock The difficulty of programming contests increases.
\newblock In \emph{International Conference on Informatics in Secondary
  Schools-Evolution and Perspectives}, pages 72--85. Springer.

\bibitem[{Gertin(2012)}]{gertin2012maximizing}
Thomas Gertin. 2012.
\newblock \emph{Maximizing the cost of shortest paths between facilities
  through optimal product category locations}.
\newblock Ph.D. thesis.

\bibitem[{Glorot and Bengio(2010)}]{Glorot10understandingthe}
Xavier Glorot and Yoshua Bengio. 2010.
\newblock Understanding the difficulty of training deep feedforward neural
  networks.
\newblock In \emph{In Proceedings of the International Conference on Artificial
  Intelligence and Statistics (AISTATS’10). Society for Artificial
  Intelligence and Statistics}.

\bibitem[{Glorot et~al.(2011)Glorot, Bordes, and Bengio}]{glorot2011domain}
Xavier Glorot, Antoine Bordes, and Yoshua Bengio. 2011.
\newblock Domain adaptation for large-scale sentiment classification: A deep
  learning approach.
\newblock In \emph{Proceedings of the 28th international conference on machine
  learning (ICML-11)}, pages 513--520.

\bibitem[{Gulwani et~al.(2017)Gulwani, Polozov, Singh
  et~al.}]{gulwani2017program}
Sumit Gulwani, Oleksandr Polozov, Rishabh Singh, et~al. 2017.
\newblock Program synthesis.
\newblock \emph{Foundations and Trends{\textregistered} in Programming
  Languages}, 4(1-2):1--119.

\bibitem[{Hansen and Salamon(1990)}]{58871}
L.~K. Hansen and P.~Salamon. 1990.
\newblock \href {https://doi.org/10.1109/34.58871} {Neural network ensembles}.
\newblock \emph{IEEE Transactions on Pattern Analysis and Machine
  Intelligence}, 12(10):993--1001.

\bibitem[{Hearst et~al.(1998)Hearst, Dumais, Osuna, Platt, and
  Scholkopf}]{hearst1998support}
Marti~A. Hearst, Susan~T Dumais, Edgar Osuna, John Platt, and Bernhard
  Scholkopf. 1998.
\newblock Support vector machines.
\newblock \emph{IEEE Intelligent Systems and their applications}, 13(4):18--28.

\bibitem[{Honnibal and Montani(2017)}]{spacy2}
Matthew Honnibal and Ines Montani. 2017.
\newblock spacy 2: Natural language understanding with bloom embeddings,
  convolutional neural networks and incremental parsing.
\newblock \emph{To appear}.

\bibitem[{Hopkins et~al.(2017)Hopkins, Petrescu-Prahova, Levin, Le~Bras,
  Herrasti, and Joshi}]{hopkins2017beyond}
Mark Hopkins, Cristian Petrescu-Prahova, Roie Levin, Ronan Le~Bras, Alvaro
  Herrasti, and Vidur Joshi. 2017.
\newblock Beyond sentential semantic parsing: Tackling the math sat with a
  cascade of tree transducers.
\newblock In \emph{Proceedings of the 2017 Conference on Empirical Methods in
  Natural Language Processing}, pages 795--804.

\bibitem[{Hosseini et~al.(2014)Hosseini, Hajishirzi, Etzioni, and
  Kushman}]{hosseini2014learning}
Mohammad~Javad Hosseini, Hannaneh Hajishirzi, Oren Etzioni, and Nate Kushman.
  2014.
\newblock Learning to solve arithmetic word problems with verb categorization.
\newblock In \emph{Proceedings of the 2014 Conference on Empirical Methods in
  Natural Language Processing (EMNLP)}, pages 523--533.

\bibitem[{Howard and Ruder(2018)}]{howard2018universal}
Jeremy Howard and Sebastian Ruder. 2018.
\newblock Universal language model fine-tuning for text classification.
\newblock In \emph{Proceedings of the 56th Annual Meeting of the Association
  for Computational Linguistics (Volume 1: Long Papers)}, volume~1, pages
  328--339.

\bibitem[{Iyer et~al.(2016)Iyer, Konstas, Cheung, and
  Zettlemoyer}]{iyer2016summarizing}
Srinivasan Iyer, Ioannis Konstas, Alvin Cheung, and Luke Zettlemoyer. 2016.
\newblock Summarizing source code using a neural attention model.
\newblock In \emph{Proceedings of the 54th Annual Meeting of the Association
  for Computational Linguistics (Volume 1: Long Papers)}, volume~1, pages
  2073--2083.

\bibitem[{Joulin et~al.(2016)Joulin, Grave, Bojanowski, and
  Mikolov}]{joulin2016bag}
Armand Joulin, Edouard Grave, Piotr Bojanowski, and Tomas Mikolov. 2016.
\newblock Bag of tricks for efficient text classification.
\newblock \emph{arXiv preprint arXiv:1607.01759}.

\bibitem[{Kim(2014)}]{kim2014convolutional}
Yoon Kim. 2014.
\newblock Convolutional neural networks for sentence classification.
\newblock \emph{arXiv preprint arXiv:1408.5882}.

\bibitem[{Kingma and Ba(2014)}]{kingma2014adam}
Diederik~P Kingma and Jimmy Ba. 2014.
\newblock Adam: A method for stochastic optimization.
\newblock \emph{arXiv preprint arXiv:1412.6980}.

\bibitem[{Kurata et~al.(2016)Kurata, Xiang, and Zhou}]{kurata2016improved}
Gakuto Kurata, Bing Xiang, and Bowen Zhou. 2016.
\newblock Improved neural network-based multi-label classification with better
  initialization leveraging label co-occurrence.
\newblock In \emph{Proceedings of the 2016 Conference of the North American
  Chapter of the Association for Computational Linguistics: Human Language
  Technologies}, pages 521--526.

\bibitem[{Kushman et~al.(2014)Kushman, Artzi, Zettlemoyer, and
  Barzilay}]{kushman2014learning}
Nate Kushman, Yoav Artzi, Luke Zettlemoyer, and Regina Barzilay. 2014.
\newblock Learning to automatically solve algebra word problems.
\newblock In \emph{Proceedings of the 52nd Annual Meeting of the Association
  for Computational Linguistics (Volume 1: Long Papers)}, volume~1, pages
  271--281.

\bibitem[{Lai et~al.(2015)Lai, Xu, Liu, and Zhao}]{lai2015recurrent}
Siwei Lai, Liheng Xu, Kang Liu, and Jun Zhao. 2015.
\newblock Recurrent convolutional neural networks for text classification.
\newblock In \emph{AAAI}, volume 333, pages 2267--2273.

\bibitem[{Lin et~al.(2017)Lin, Wang, Pang, Vu, Zettlemoyer, and
  Ernst}]{lin2017program}
Xi~Victoria Lin, Chenglong Wang, Deric Pang, Kevin Vu, Luke Zettlemoyer, and
  Michael~D Ernst. 2017.
\newblock Program synthesis from natural language using recurrent neural
  networks.
\newblock \emph{University of Washington Department of Computer Science and
  Engineering, Seattle, WA, USA, Tech. Rep. UW-CSE-17-03-01}.

\bibitem[{Liu et~al.(2017)Liu, Chang, Wu, and Yang}]{liu2017deep}
Jingzhou Liu, Wei-Cheng Chang, Yuexin Wu, and Yiming Yang. 2017.
\newblock Deep learning for extreme multi-label text classification.
\newblock In \emph{Proceedings of the 40th International ACM SIGIR Conference
  on Research and Development in Information Retrieval}, pages 115--124. ACM.

\bibitem[{Matsuzaki et~al.(2017)Matsuzaki, Ito, Iwane, Anai, and
  Arai}]{matsuzaki2017semantic}
Takuya Matsuzaki, Takumi Ito, Hidenao Iwane, Hirokazu Anai, and Noriko~H Arai.
  2017.
\newblock Semantic parsing of pre-university math problems.
\newblock In \emph{Proceedings of the 55th Annual Meeting of the Association
  for Computational Linguistics (Volume 1: Long Papers)}, volume~1, pages
  2131--2141.

\bibitem[{McDowell(2016)}]{mcdowell2016cracking}
Gayle~Laakmann McDowell. 2016.
\newblock \emph{Cracking the Coding Interview: 189 Programming Questions and
  Solutions}.
\newblock CareerCup, LLC.

\bibitem[{Mehta et~al.(2017)Mehta, Mishra, Athavale, Shrivastava, and
  Sharma}]{mehta2017deep}
Purvanshi Mehta, Pruthwik Mishra, Vinayak Athavale, Manish Shrivastava, and
  Dipti Sharma. 2017.
\newblock Deep neural network based system for solving arithmetic word
  problems.
\newblock \emph{Proceedings of the IJCNLP 2017, System Demonstrations}, pages
  65--68.

\bibitem[{Nam et~al.(2014)Nam, Kim, Menc{\'\i}a, Gurevych, and
  F{\"u}rnkranz}]{nam2014large}
Jinseok Nam, Jungi Kim, Eneldo~Loza Menc{\'\i}a, Iryna Gurevych, and Johannes
  F{\"u}rnkranz. 2014.
\newblock Large-scale multi-label text classification—revisiting neural
  networks.
\newblock In \emph{Joint european conference on machine learning and knowledge
  discovery in databases}, pages 437--452. Springer.

\bibitem[{Paszke et~al.(2017)Paszke, Gross, Chintala, Chanan, Yang, DeVito,
  Lin, Desmaison, Antiga, and Lerer}]{paszke2017automatic}
Adam Paszke, Sam Gross, Soumith Chintala, Gregory Chanan, Edward Yang, Zachary
  DeVito, Zeming Lin, Alban Desmaison, Luca Antiga, and Adam Lerer. 2017.
\newblock Automatic differentiation in pytorch.
\newblock In \emph{NIPS-W}.

\bibitem[{Pedregosa et~al.(2011)Pedregosa, Varoquaux, Gramfort, Michel,
  Thirion, Grisel, Blondel, Prettenhofer, Weiss, Dubourg, Vanderplas, Passos,
  Cournapeau, Brucher, Perrot, and Duchesnay}]{scikit-learn}
F.~Pedregosa, G.~Varoquaux, A.~Gramfort, V.~Michel, B.~Thirion, O.~Grisel,
  M.~Blondel, P.~Prettenhofer, R.~Weiss, V.~Dubourg, J.~Vanderplas, A.~Passos,
  D.~Cournapeau, M.~Brucher, M.~Perrot, and E.~Duchesnay. 2011.
\newblock Scikit-learn: Machine learning in {P}ython.
\newblock \emph{Journal of Machine Learning Research}, 12:2825--2830.

\bibitem[{Pennington et~al.(2014)Pennington, Socher, and
  Manning}]{Pennington14glove:global}
Jeffrey Pennington, Richard Socher, and Christopher~D. Manning. 2014.
\newblock Glove: Global vectors for word representation.
\newblock In \emph{In EMNLP}.

\bibitem[{Rumelhart et~al.(1986)Rumelhart, Hinton, and
  Williams}]{rumelhart:errorpropnonote}
David~E. Rumelhart, Geoffrey~E. Hinton, and Ronald~J. Williams. 1986.
\newblock Learning internal representations by error propagation.
\newblock In David~E. Rumelhart and James~L. Mcclelland, editors,
  \emph{Parallel Distributed Processing: Explorations in the Microstructure of
  Cognition, {V}olume 1: {F}oundations}, pages 318--362. MIT Press, Cambridge,
  MA.

\bibitem[{Sudha et~al.(2017)Sudha, Kumar, Nagappan, and
  Suresh}]{sudha2017classification}
S~Sudha, A~Arun Kumar, M~Muthu Nagappan, and R~Suresh. 2017.
\newblock Classification and recommendation of competitive programming problems
  using cnn.
\newblock In \emph{International Conference on Intelligent Information
  Technologies}, pages 262--272. Springer.

\bibitem[{Wang and Manning(2012)}]{Wang:2012:BBS:2390665.2390688}
Sida Wang and Christopher~D. Manning. 2012.
\newblock \href {http://dl.acm.org/citation.cfm?id=2390665.2390688} {Baselines
  and bigrams: Simple, good sentiment and topic classification}.
\newblock In \emph{Proceedings of the 50th Annual Meeting of the Association
  for Computational Linguistics: Short Papers - Volume 2}, ACL '12, pages
  90--94, Stroudsburg, PA, USA. Association for Computational Linguistics.

\bibitem[{Wang et~al.(2017)Wang, Liu, and Shi}]{wang2017deep}
Yan Wang, Xiaojiang Liu, and Shuming Shi. 2017.
\newblock Deep neural solver for math word problems.
\newblock In \emph{Proceedings of the 2017 Conference on Empirical Methods in
  Natural Language Processing}, pages 845--854.

\bibitem[{Yang et~al.(2016)Yang, Yang, Dyer, He, Smola, and
  Hovy}]{yang2016hierarchical}
Zichao Yang, Diyi Yang, Chris Dyer, Xiaodong He, Alex Smola, and Eduard Hovy.
  2016.
\newblock Hierarchical attention networks for document classification.
\newblock In \emph{Proceedings of the 2016 Conference of the North American
  Chapter of the Association for Computational Linguistics: Human Language
  Technologies}, pages 1480--1489.

\bibitem[{Zhong et~al.(2017)Zhong, Xiong, and Socher}]{zhong2017seq2sql}
Victor Zhong, Caiming Xiong, and Richard Socher. 2017.
\newblock Seq2sql: Generating structured queries from natural language using
  reinforcement learning.
\newblock \emph{arXiv preprint arXiv:1709.00103}.

\end{thebibliography}
\bibliographystyle{acl_natbib}

\appendix

\section{Appendix}
\subsection{Experiments with limited training data}
\begin{figure}
  \begin{minipage}{0.48\textwidth}
  \centering
      \includegraphics[scale=.35]{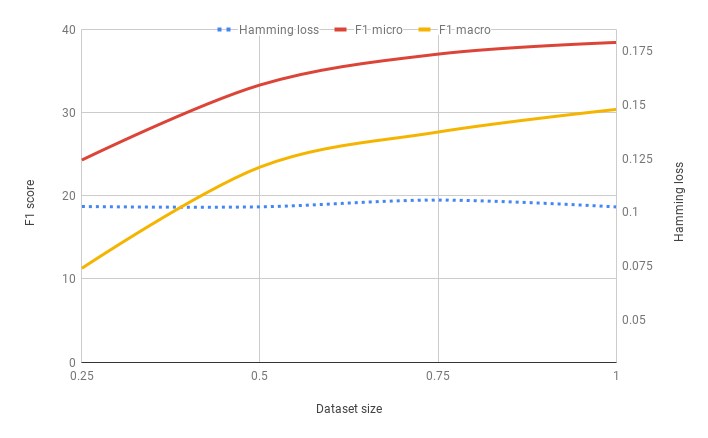}
      \caption{F1 micro, macro and hamming loss variation when models trained on percentage of the CFMC20 dataset.  Note that scale for both the F1 scores is given on the left and the one for hamming loss is given on the right.}
      \label{fig:datasetsize}
  \end{minipage}
\end{figure}

We wanted to see how the dataset size affects the performance of the classifier. So, we train a CNN classifier on 25, 50, 75, and 100 percent of the CFML20 dataset. As expected, we find that the performance of the classifier improves with increase in size of the training data. The F1 micro and macro scores increase, and the hamming loss decreases. For the F1 scores, higher is better, while for hamming loss lower is better. See figure \ref{fig:datasetsize}.

\subsection{Evaluation Metrics}

\subsection{Multiclass: Accuracy} Accuracy is the percentage of labels correctly predicted. Note that for multiclass classification the micro-averaged F1 score is equal to the accuracy.
\subsection{Multiclass: Macro-averaged F1 score} 
Macro-averaged F1 score is computed by first computing the F1 score for each class independently and then take an averaging all the F1 scores. This metric treats all the classes as equal, independent of their frequency in the test set.
\subsection{Multiclass: Weighted macro-averaged F1 score} 
Weighted macro-averaged F1 score is computed by first computing the F1 score for each class independently and then take an averaging all the F1 scores, weighted by their support.
\subsection{Multilabel: Hamming loss} Hamming loss is the proportion of mis-classified examples in the dataset.
\subsection{Multilabel: Micro-averaged F1 score} It is the F-measure averaging on the prediction matrix. The individual true positives, false positives, and false negatives are summed up across labels/classes and then the F-measure is calculated.
\subsection{Multilabel: Macro-averaged F1 score} Macro-averaged F1 score is calculated by computing the F1 score for each of the labels, then averaging the label wise F1 scores.

\section{Human accuracy}
We did a human study with 5 participants on the CFML20 dataset \ref{table:human}. Each participant is a recent graduate in computer science and is a frequent competitive programmer. You can see the results in \ref{table:human}
\begin{table}[t!]
\begin{tabularx}{\textwidth}{|l|l|l|}
 \cline{1-3}  \multirow{2}{*}{\bf Classifier}
      
          & \multicolumn{2}{c|}{\bf 20multi subset} \\ \cline{2-3}             
  & F1 micro & F1 macro \\ 
  \cline{1-3}
Human 1 & 56.3 & 42.3  \\
Human 2 & 46.1 & 38.7  \\
Human 3 & 51.1 & 40.6  \\
Human 4 & 48.4 & 42.8  \\
Human 5 & 57.3 & 49.1  \\
Human Average & 51.8 & 42.7  \\
\cline{1-3}
\end{tabularx}  
\caption{Human accuracy on a 100 sized subset of the CFML20 dataset. HL is the hamming loss.}\label{table:human}
\end{table}

\section{Classes classification in CFML20}
\subsection{Problem category}
Following classes belong to Problem category:
probabilities, 
geometry, 
combinatorics, 
number theory, 
strings, 
trees, 
graphs, 
math, 
data structures 

\subsection{Solution Category}
Following classes belong to Solution category:
dsu, 
binary search, 
dfs and similar, 
constructive algorithms, 
brute force, 
greedy, 
dp, 
bitmask, 
two pointers, 
sortings, 
implementation 
\end{document}